# Co-Arg: Cogent Argumentation with Crowd Elicitation

**Mihai Boicu, Dorin Marcu, Gheorghe Tecuci, Lou Kaiser, Chirag Uttamsingh, Navya Kalale**

Learning Agents Center, Volgenau School of Engineering, George Mason University, Fairfax, VA 22030, USA
{mboicu; dmarcu; tecuci; lkaiser4; cuttamsi; nkalale}@gmu.edu

**Abstract**

This paper presents Co-Arg, a new type of cognitive assistant to an intelligence analyst that enables the synergistic integration of analyst imagination and expertise, computer knowledge and critical reasoning, and crowd wisdom, to draw defensible and persuasive conclusions from masses of evidence of all types, in a world that is changing all the time. Co-Arg's goal is to improve the quality of the analytic results and enhance their understandability for both experts and novices. The performed analysis is based on a sound and transparent argumentation that links evidence to conclusions in a way that shows very clearly how the conclusions have been reached, what evidence was used and how, what is not known, and what assumptions have been made. The analytic results are presented in a report describes the analytic conclusion and its probability, the main favoring and disfavoring arguments, the justification of the key judgments and assumptions, and the missing information that might increase the accuracy of the solution.

## Introduction

In January 2017, the Intelligence Advanced Research Projects Activity launched the Crowdsourcing Evidence, Argumentation, Thinking and Evaluation (CREATE) program, a 4.5-year effort to develop and experimentally test systems that use crowdsourcing and structured analytic techniques to improve analytic reasoning (IARPA, 2016).

There are four performers in CREATE, each led by a university (George Mason University, Monash University, Syracuse University, and University of Melbourne), and each investigating a different approach.

George Mason University is developing Co-Arg, a cognitive assistant based on a theory of evidence-based reasoning with Wigmorean inference networks (Schum, 1987; 2001; Tecuci et al., 2016a, 2016b). Co-Arg supports answering intelligence questions by synergistically integrating analyst imagination and expertise, computer knowledge and critical reasoning, and crowd wisdom.

Monash University is developing BARD (Bayesian ARgumentation via Delphi), a system that answers intelligence questions by using causal Bayesian networks as underlying structured representations for argument analysis and automated Delphi methods to bring groups of analysts to a consensus analysis.

Syracuse University is developing TRACE (Trackable Reasoning and Analysis for Crowdsourcing and Evaluation), the goal of which is to experimentally evaluate existing structured analytic techniques in order to determine the most effective ones.

Finally, University of Melbourne is developing SWARM (Smartly-assembled Wiki-style Argument Marshalling), an online collaboration platform supporting evidence-based reasoning by cultivating user engagement, exploiting natural expertise, and supporting rich collaboration.

There are also two control systems based on GoogleDocs, Conclude (for an individual user) and Concur (for a team).

All these six systems are integrated into a common evaluation environment developed by the testing and evaluation team (T&E) consisting of Good Judgement Inc. and John Hopkins University's Applied Physics Laboratory.

The systems are evaluated using volunteers who receive problems to solve and, for each problem, choose the system to use. Two of the systems, TRACE and SWARM, as well as GoogleDocs, require minimal training to use them. BARD users need to be taught Bayesian reasoning and how to build Bayesian networks with BARD. Similarly, Co-Arg users need to be taught a theory of evidence-based reasoning and to build Wigmorean networks with Co-Arg. Thus a significant challenge for these two systems is how to teach volunteers, in a couple of hours, to be both willing and able to solve complex problems with these systems.

Each performer developed a short video to encourage the volunteers to use its system. The video introducing Co-Arg is accessible at: https://youtu.be/7_fuCELpUL0

For each evaluation problem posted by T&E, each user chooses one of the six systems and is assigned to a team. Team sizes depend on the system and the available users. SWARM attempts to form teams of 24 users, BARD forms teams of 12 users, and Co-Arg forms teams of up to 12 users. TRACE users work independently. Concur uses teams of size 12 and Conclude single users.

In the following we provide an overview of the design and use of the Co-Arg system that was developed during the 2-

year long Phase 1 of the CREATE program.

## Evaluation Problems

The evaluation problems were selected from those proposed by the performers and T&E.

The GMU team developed evidence-based reasoning problems of the type intelligence analysts routinely encounter:
- Describe a situation, an intelligence question to answer, and additional imperfect information to use in answering the question.
- Require marshaling and evaluation of evidence for both credibility and relevance to the hypothesis, incorporation of logical inferences and assumptions to answer the question posed, and justification for assumptions with an assessment of their credibility.
- Provide information from different types of sources, including human sources, intercepted communications, and documentary evidence. Information includes both favoring and disfavoring evidence for different hypotheses, and need to be assessed for credibility.
- Require formulation of an answer to the intelligence question in the form of a production report.

Below are sample questions asked in these problems:
- Is John Ventura, the director of Polombia's intelligence service, involved in the drug sale to Markistan that is being arranged by Joe Salazar?
- Which surface-to-air missile system is Manada selling Sindia?
- Why did Finance Minister Goodguy resign?

## Co-Arg Overview

The use of Co-Arg in the T&E experiment is presented in Figure 1.

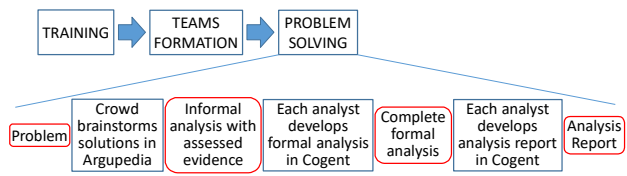

*Figure 1: Co-Arg workflow.*

First the interested users are rapidly trained in evidence-based reasoning and the use of Co-Arg. Then, as they volunteer to use Co-Arg in solving the current problem published by T&E, they are placed in teams. Each team follows the process from the bottom of Figure 1 that is described in more details in the following sections.

The overall architecture of Co-Arg, as integrated into the T&E evaluation environment, is shown in Figure 2.

Co-Arg has two complementary components: Argupedia and Cogent. Argupedia receives a problem to be solved and the entire team uses it to brainstorm possible answers to the intelligence question(s) asked. Then, for each imagined possible answer or hypothesis, the team brainstorms and collaborates in developing a brief informal argumentation.

The informal argumentations are passed to Cogent where each user, independently of the other users of the team, develops formal argumentations and an analysis report that answers the intelligence question(s) asked.

Notice that, as opposed to the other systems that produce one solution per team, Co-Arg can produce as many solutions as team members. These solutions, however, are only partially independent because team members work collaboratively for 40% of the time and independently for the remaining 60% of the time.

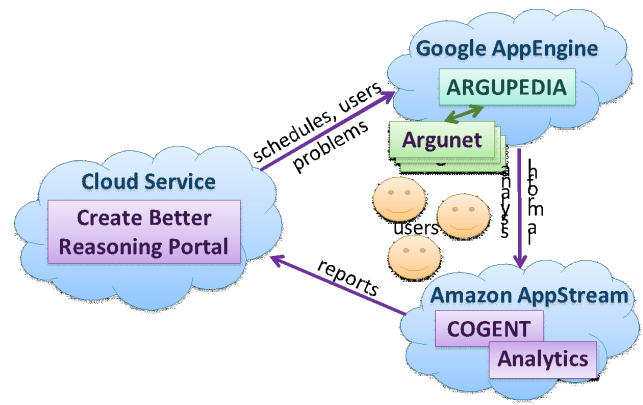

*Figure 2: Co-Arg in the evaluation environment.*

## Training in Evidence-Based Reasoning and the Use of Co-Arg

A very significant challenge for us was to develop a brief but effective training course in evidence-based reasoning and the use of Co-Arg. Part of this training is publicly accessible at: https://sites.google.com/site/learncogent/

The course consists of short video lessons and equivalent textual descriptions, as well as hands on exercises. In the following we provide a fragment of this course that will facilitate the understanding of the rest of this paper.

The first step in answering an intelligence question is to imagine possible answers. Each such answer is a hypothesis to be assessed. You need to determine which hypotheses is best supported by the available information.

To assess the likelihood that a hypothesis is true you have to develop an ***argumentation***. *An argumentation is a reasoning structure that shows how the evidence and our assumptions prove or refute the hypothesis.*

Consider, for example, the hypothesis "Hakka has chemical weapons." One way to prove it is to show that Hakka, which is an apocalyptic sect, develops chemical weapons. This, in turn, might be proved by showing that

Hakka has the necessary expertise, production materials, and funds (see top part of Figure 3). Each of these sub-hypotheses might be proved based on the available information or by making assumptions.

Consider the following information: "A source, who has reported accurately in the past, indicated that Hakka has a member with a bachelor's degree in chemistry." This information supports the truthfulness of the hypothesis that "Hakka has expertise to develop chemical weapons," and is therefore evidence for this hypothesis. Let's name it: "E1 Chemical expert" (see Figure 3).

In general, *evidence is any item of information that favors or disfavors the truthfulness of a hypothesis*. We say that the evidence is *relevant* to that hypothesis.

A difficulty, however, is that our evidence is always **incomplete**, and may also be **inconclusive**, **ambiguous**, and **dissonant**. And it has various degrees of *credibility*. For example, we may assess that the credibility of evidence E1 is *very likely (80-95%)*.

It is also possible that, for some of the sub-hypotheses, such as "Hakka has funds," we may not have any evidence. In such a case we may treat it as an assumption. *An **assumption** is a statement taken to be true, based on knowledge about similar situations and commonsense reasoning, without having any supporting evidence.* For example, we may assume that it is *likely (55-70%)* that "Hakka has funds."

As a result of all these uncertainties, we will not be able to prove that the top hypothesis is definitely true or definitely false, but we will be able to estimate the probability of it being true or false, such as "It is *likely (55-70%) that Hakka has chemical weapons.*"

One way of improving the quality of the analysis is to collect information that would either corroborate or contradict the validity of the assumptions made.

In the following, we are going to learn how to answer questions based on imperfect information, by generating competing hypotheses and building argumentations for assessing which hypothesis is the most likely. We will start by introducing the main characteristics or credentials of evidence: *credibility*, *relevance*, and *inferential force*.

It is important to distinguish between evidence about a fact and the fact itself. Consider the item of evidence from Figure 3. Can we conclude from it that Hakka has a member with a bachelor's degree in chemistry? No. At issue here is the *credibility* of the source who may or may not be telling the truth. *Credibility of an item of evidence is the extent to which the evidence may be believed. This assessment can be influenced by many things, including doubts about the source's veracity or by more credible information that contradicts this item of evidence.*

We can assess the credibility of this item of evidence by answering the following question: *What is the probability that the evidence is true?* The source of this evidence has reported accurately in the past. We can therefore assume that this current report is very likely to be true.

Another credential or property of evidence is its relevance. *The **relevance** of an item of evidence indicates how **strongly** this item **supports a specific** hypothesis in the argument.* Relevance depends on how **recent** the evidence is, how **unambiguous** it is, and how **conclusive** the link between the evidence and the hypothesis is. The evidence may be unambiguous but it may support more than one hypothesis.

We can assess the relevance by answering the question: *Assuming that the evidence is true, what is the probability that the hypothesis is true?*

When we have evidence about a fact and the hypothesis is the fact itself, the relevance of evidence is certain. Indeed, if we assume that the evidence is true, then the hypothesis is true (see the bottom part of Figure 4).

But let us consider the hypothesis that "Hakka has expertise to develop chemical weapons." If Hakka has a member with a bachelor's degree in chemistry, what is the probability that it has expertise to develop chemical weapons? A bachelor program in chemistry does provide the basic knowledge for chemical weapons development, but this does not necessarily prove that Hakka has the expertise. Indeed, the Hakka member may not have developed this expertise. Thus this item of evidence is not conclusive and we assess its relevance only as *likely* (see top part of Figure 4). Such explanations for assessments of relevance that are less than certain will clarify the reasoning and can be used in the completed argumentation to support the conclusion.

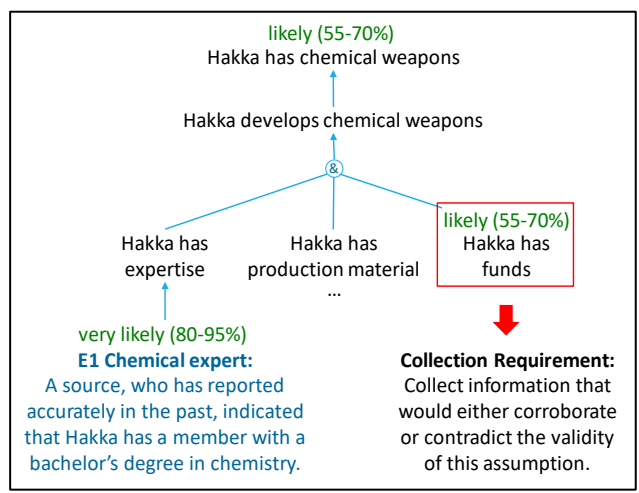

*Figure 3: Simple argumentation.*

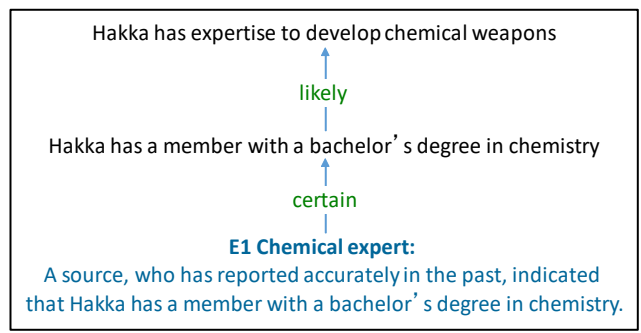

*Figure 4: Relevance of evidence and argument.*

When developing an argumentation, it is a good practice to consider, for each item of evidence, which is the corresponding fact, and then reason from that fact to the upper-level hypotheses, as illustrated in Figure 4.

The third credential of evidence is its *inferential force*. Consider the inference from the bottom part of Figure 5. We have assessed the relevance of the item of evidence E1 as *certain* because it is an inference from evidence about a fact to the fact itself. We have also assessed the credibility of E1 as *very likely* because the source has reported accurately in the past. *Inferential force* answers the question: *What is the probability of the hypothesis above based only on this item of evidence below?* In our example, the relevance of E1 is *certain*, but its credibility is only *very likely*. Therefore the probability that "Hakka has a member with a bachelor's degree in chemistry" is only *very likely*.

In general, the inferential force of an item of evidence is determined as the smaller between its credibility and its relevance. Indeed, an item of evidence that is not credible would not convince us that the hypothesis is true, no matter how relevant the provided information is. Therefore the inferential force in this circumstance would be low. Similarly, it is not enough for the item of evidence to be credible, if the information provided is not relevant to the hypothesis. The inferential force will be high only if the evidence item is both highly relevant and credible.

In this case, because we have only one item of evidence, the probability of the hypothesis "Hakka has a member with a bachelor's degree in chemistry" is given by the inferential force of this item of evidence. However, if we have more items of evidence, some favoring the truthfulness of the hypothesis, and some disfavoring it, then the probability of the hypothesis will result from the combined inferential force of all these items of evidence.

As another example, let's now consider the upper-level hypothesis from Figure 5: "Hakka has expertise to develop chemical weapons." The relevance of "Hakka has a member with a bachelor's degree in chemistry" to this hypothesis was assessed as *likely*. Because the probability of the sub-hypothesis is *very likely*, its inferential force to the top hypothesis is *likely*, the minimum of its relevance and probability.

In this case we have just this one reason and one argument for the top hypothesis to be true. Therefore the probability of the top hypothesis is the same as the inferential force of this reason. In general, however, we may have multiple arguments, some favoring the truthfulness of the top hypothesis and some disfavoring it. In such a case the probability of the top hypothesis will be given by the combined inferential force of all these arguments.

Co-Arg is based on an intuitive and easy to use system of **Baconian probabilities** (Cohen, 1977; 1989) with **Fuzzy qualifiers** (Zadeh, 1983; Negoita and Ralescu, 1975), governed by the min-max probability combination rules for conjunctions and disjunctions (Tecuci et al., 2016a, b). As was illustrated above, Co-Arg uses a positive probability scale that is a refinement of the scale provided in the Intelligence Community Directive 203 (ICD, 203): *lacking support (0-50%) < barely likely (50-55%) < likely (55-70%) < more than likely (70-80%) < very likely (80-95%) < almost certain (95-99%) < certain (100%)*.

## Team Formation

In a collaborative environment team formation is very important because the team dynamics and the quality of the work performed is dependent on the team size, the domain of expertise, relative skill levels of the group members, and the group's chemistry. We have experimented with two team formation strategies.

In our own internal evaluations of Co-Arg, the teams are randomly formed and they remain unchanged during the entire experiment. The team size is between 3 and 6 members, based on the available number of participants in each experiment. This method guarantees a random distribution of the participants.

In the T&E evaluation, for each problem, the participants select a system to use. When a participant decides to use Co-Arg, he or she is included into an ad-hoc team. If there is no open team, a new team is created and the participant is registered to that team. Once created, a team will remain open for six hours or until 12 participants join the team. If after 6 hours the team has fewer than 6 participants, the team will remain open for 12 hours or until 6 participants join the team. Depending on the total number of participants selecting Co-Arg, this team formation method may lead potential problems. For instance, if there are few participants selecting Co-Arg we may end-up with small teams, or even one-person teams. Also, it is possible that the more enthusiastic users of Co-Arg will select the system early, so we may detect a better performance for the early teams as compared to the late teams.

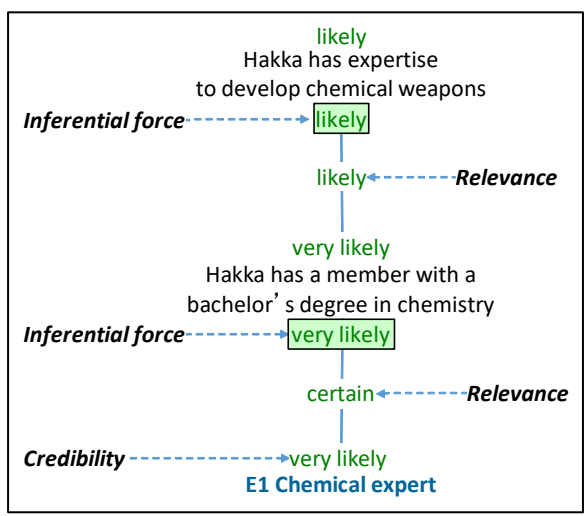

*Figure 5: Credentials of evidence and arguments.*

## Brainstorming of the Informal Analysis

At the beginning of the analysis process a participant is instructed to read the description of the problem and to imagine possible hypotheses, arguments to support or refute them, and what information supports these arguments. We introduced this step following various internal tests in which some participants complained about the framing bias (i.e., a member of the team will be influenced by how the members that already started the brainstorming, framed the problem).

Once familiarized with the problem, and having a personal opinion about the solution, the participants may start the asynchronous brainstorming process.

In the following illustration we consider the Cesium problem: It was reported that a canister containing cesium-137 is missing from the XYZ Company in MD and the intelligence question is: What happened to it.

Because the question is provided, the first task for the participants is to provide possible answers or hypotheses. In this illustration we will assume a team of 3 participants, P1, P2 and P3, with P1 being the first to formulate the answers:

*What happened to the cesium-137 canister?*
   Was stolen
   Was misplaced
   Was lost

Next P1 continues with the formulation of informal arguments to support or refute each of these hypotheses. However, in this example, we focus on the contributions of the other two members to hypotheses generation step.

Next to work on the analysis is P2. He reads the answers provided by P1 and has the following options:

- Reformulate an existing answer
- Vote for the formulation of an existing answer
- Provide a new answer
- Reject an existing answer

P2 decides to reformulate two of the answers. He also rejects the third answer, justifies the rejection, and provides another answer:

*What happened to the cesium-137 canister?*
   P2: The cesium-137 canister was stolen (1 vote)
   Team version: Was stolen (1 vote)
   P2: The cesium-137 canister was misplaced (1 vote)
   Team version: Was misplaced (1 vote)
   P2: The cesium-137 canister is being used in a project of the XYZ Company without having been checked out from XYZ warehouse

After that P2 reviews and revises the analysis done by P1 for the first two hypotheses and provides an analysis for the new hypothesis proposed by him.

P3 reads the hypotheses proposed by P1 and P2 and votes for the reformulations proposed by P2. After reading the note from P2, P3 agrees that the "Was lost" hypothesis is covered by the proposed answers. The result is the following one:

*What happened to the cesium-137 canister?*
   Team version: The cesium-137 canister was stolen (2 votes: P2, P3)
   P1: Was stolen (1 vote: P1)
   Team version: The cesium-137 canister was misplaced (2 votes: P2, P3)
   P1: Was misplaced (1 vote: P1)
   Team version: Was lost (1 vote: P1)
   Team version: The cesium-137 canister is being used in a project of the XYZ Company without having been checked out from XYZ warehouse. (2 votes: P2, P3)

Because P2 and P3 have modified the hypotheses, they are marked as incomplete for P1. Therefore, the next time P1 logs in, the system guides her to review the modifications proposed by P2 and P3. Let us assume that P1 also votes for these modifications. As a result, the initial formulations of the hypotheses remain without any vote and are deleted.

The same process is used for the follow-on brainstorming task of providing informal arguments for each of the hypotheses. For example, the informal argument developed by the team for the hypothesis "The cesium-137 canister was stolen," is the following one:

*A truck entered the company, the canister was stolen from the locker, loaded into the truck, and the truck left with the canister.*

After brainstorming informal arguments, the participants associate favoring and disfavoring evidence to each of these arguments.

The last brainstorming phase is to assess the credibility of the evidence used. Each participant is asked to assess the credibility of each item of evidence, and these individual assessments are combined into a team assessment. For example, E1 was assessed as *likely (55-70%)* by P1, as *more than likely (70-80%)* by P2, and as *barely likely (50-55%)* by P3, resulting in a team credibility of *likely (55-70%)*.

The final informal argumentation for "The cesium-137 canister was stolen" is shown in Figure 6.

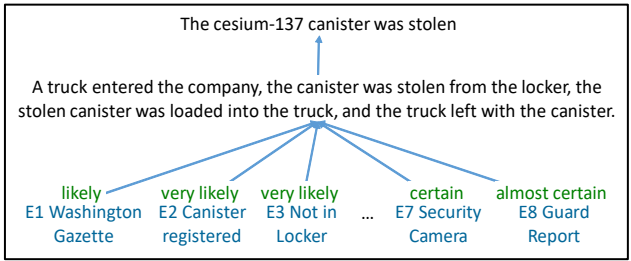

*Figure 6: Informal argumentation.*

To summarize, the participants first brainstorm the reformulation of each intelligence question, and then its possible answers (i.e., hypotheses). After that they brainstorm informal arguments for each hypothesis, relevant evidence for each argument, and credibility for each evidence item. They have two means to perform the brainstorming: ***problem check-list*** (a guided step by step process, based on a list of tasks that need to be completed in a predefined order), and ***graphical analysis*** (a graphical way providing the freedom to perform the tasks in any order).

## Formal Analysis

The informal analysis developed in Argupedia is imported into Cogent by each member of the team who continues to work independently to develop his or her own formal analysis. For example, guided by the informal argumentation in Figure 6, a user has developed the formal argumentation from Figure 7. This is generally done by decomposing the top hypothesis into simpler and simpler hypotheses, down to the level of simple hypotheses that are assessed based on the relevant evidence. At this point the disfavoring arguments and evidence are also inserted into the developed argumentation.

Figure 8 shows the formal analysis with the three alternative hypotheses and the top parts of their argumentations.

Once the formal argumentations for all the hypotheses are developed, the user invokes the Analytics Assistant that checks the argumentations for errors and warnings. For example, Figure 9 shows some of the warning messages.

Notice that the user is alerted that he may be biased toward confirming the hypothesis "The cesium-137 canister was stolen" because this hypothesis has only favoring

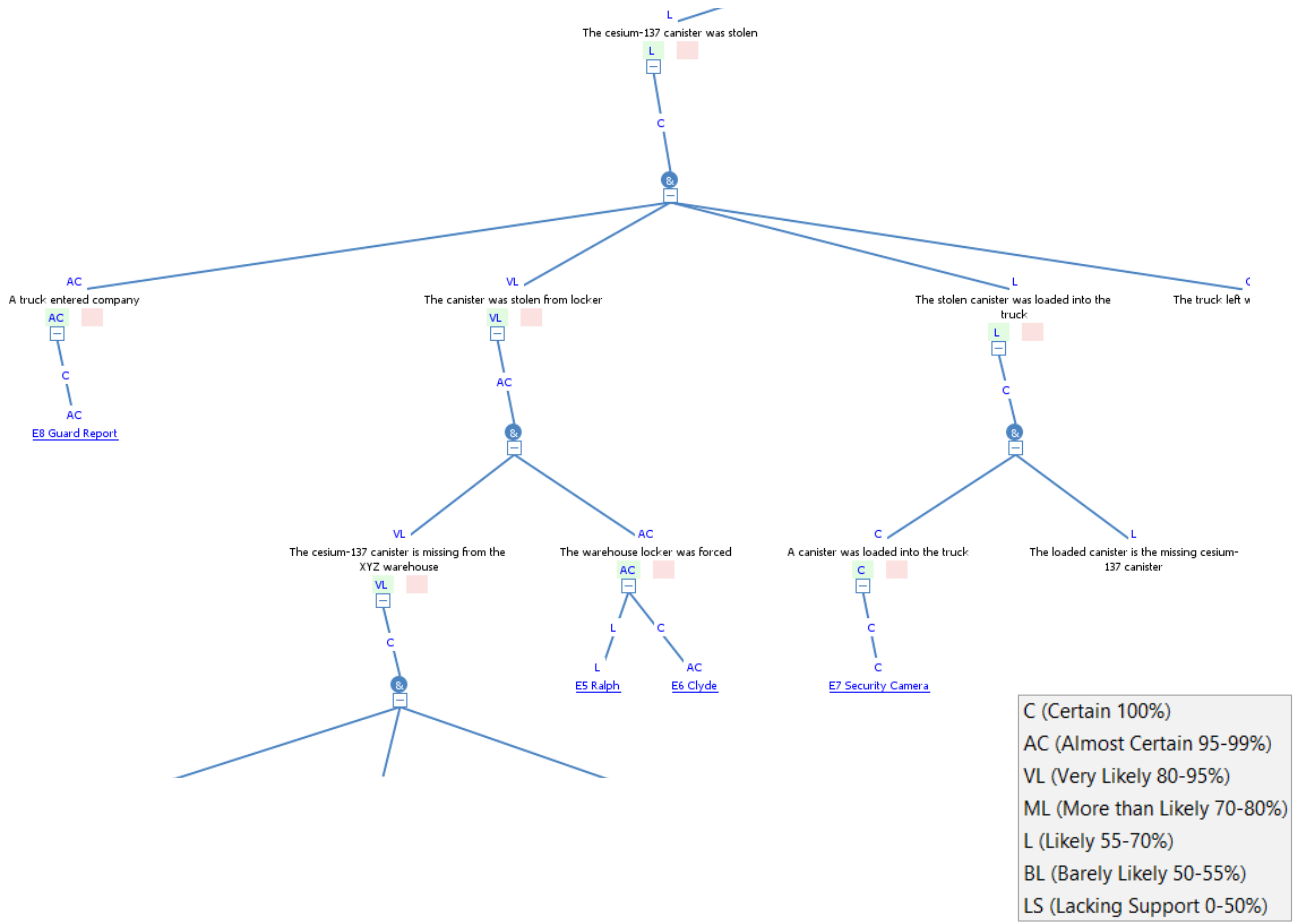

*Figure 7: Formal argumentation.*

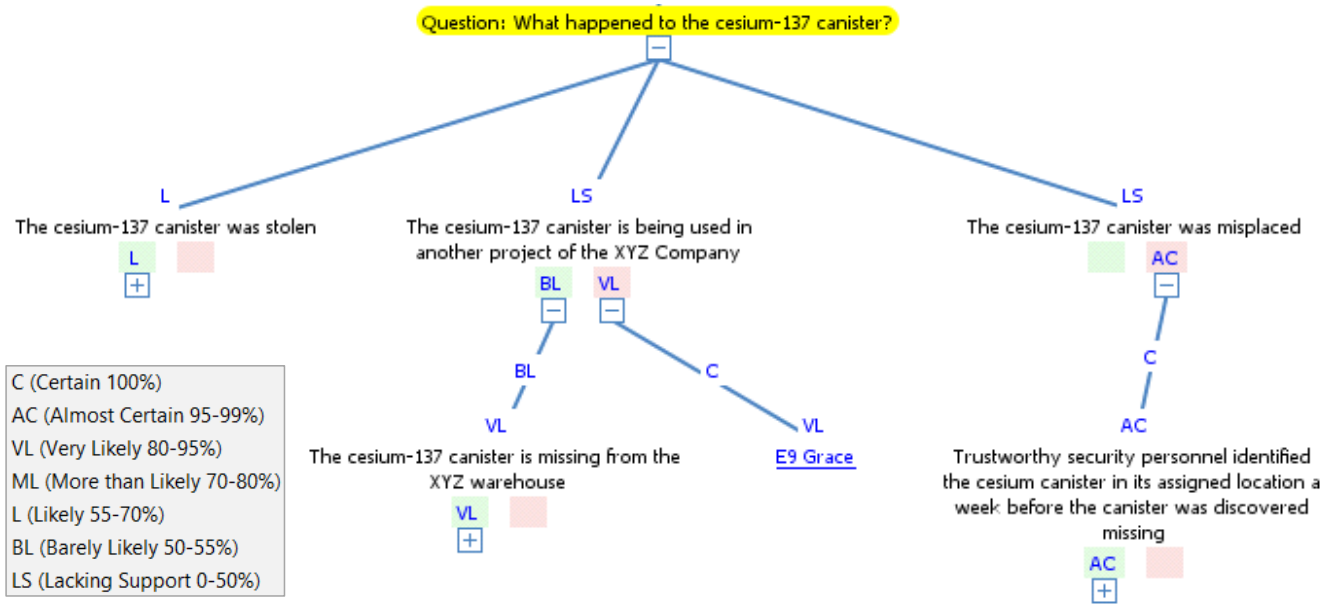

*Figure 8: Formal analysis.*

arguments and each such argument has only favoring evidence.

Cogent may also detect potential **Satisficing bias** (choosing the first hypothesis that appears good enough, rather than carefully identifying all possible hypotheses and determining which one is the most consistent with the evidence), as well as potential **Absence of Evidence bias** (failure to consider the degree of completeness of the available evidence). Many other biases are avoided because explicit argumentations are developed and the system employs an intuitive system of symbolic probabilities.

*Figure 9: Messages from the Analytics Assistant.*

## Report Development

After the user is satisfied with the developed argumentation, he or she develops a production report which is the final solution assessed by T&E.

The Report Assistant automatically generates a structured report that corresponds to the developed argumentation. The user, however, may transform it into a production report by editing it to be more comprehensible and persuasive. This also includes inserting fragments of the argumentation.

The top part of the production report is shown in Figure 10. The links in the report are to fragments of the argumentation and evidence included in the report's appendix.

*Figure 10: Fragment of the production report.*

The Cogent component of Co-Arg (Tecuci et al., 2018) is the latest in a sequence of increasingly more practical cognitive assistants for intelligence-analysis: Disciple-LTA (Tecuci et al., 2005; 2007; 2008), TIACRITIS (Tecuci et al., 2011), Disciple-CD (Tecuci et al., 2016b), and a previous version of Cogent (Tecuci et al., 2015).

## Quality of Reasoning Assessment

The report is assessed for its quality of reasoning by using the following criteria: (1) Hypotheses generation and accuracy of solution; (2) Argumentation structure and reasoning; (3) Identification of sources and assessment of credibility of evidence; and (4) Identification of key missing information and assumptions.

## Conclusions

The following summarizes some of the advantages of using Co-Arg to answer intelligence questions:
- People work in different locations at different times, making collaboration difficult. Co-Arg enables asynchronous collaboration between them.
- There is an overwhelming amount of information and trying to connect all the dots in your head is not possible. Co-Arg helps make defensible and persuasive arguments, based on all the available information.
- Humans tend to overlook arguments and evidence that do not conform to built-in biases. Co-Arg allows you to rigorously consider both favoring and disfavoring arguments and evidence.
- Direct evidence for a hypothesis may not be available. Co-Arg allows you to use circumstantial evidence and assumptions.
- Things are not always what they appear to be. People lie. They also exaggerate and have biases they are unaware of. Co-Arg allows you to take into account the credibility of evidence.
- New information comes to us in real time. Co-Arg allows you to integrate this information in your analysis without starting over or over-reacting.

## Acknowledgements

This research is based upon work supported in part by the Office of the Director of National Intelligence (ODNI), Intelligence Advanced Research Projects Activity (IARPA) under contract number 2017-16112300009, and by George Mason University. The views and conclusions contained herein are those of the authors and should not be interpreted as necessarily representing the official policies, either expressed or implied, of ODNI, IARPA, or the U.S. Government. The U.S. Government is authorized to reproduce and distribute reprints for governmental purposes notwithstanding any copyright annotation therein.